\documentclass[10pt,twocolumn,letterpaper]{article}

\usepackage{iccv}       

\usepackage{amsmath,amsfonts,bm}

\def\eqref#1{equation~\ref{#1}}

\def\1{\bm{1}}

\DeclareMathAlphabet{\mathsfit}{\encodingdefault}{\sfdefault}{m}{sl}
\SetMathAlphabet{\mathsfit}{bold}{\encodingdefault}{\sfdefault}{bx}{n}

\DeclareMathOperator*{\argmin}{arg\,min}

\definecolor{cvprblue}{rgb}{0.21,0.49,0.74}
\usepackage[pagebackref,breaklinks,colorlinks,allcolors=cvprblue]{hyperref}

\def\paperID{8725} 
\def\confName{ICCV}
\def\confYear{2025}

\usepackage{url}
\usepackage{colortbl} 

\usepackage{amsmath}
\usepackage{amssymb} 
\usepackage[inline]{enumitem}
\usepackage{algorithm,algpseudocode}
\usepackage{multirow}

\usepackage[normalem]{ulem}
\useunder{\uline}{\ul}{}

\definecolor{darkpowderblue}{rgb}{0.0, 0.2, 0.6}
\definecolor{bittersweet}{rgb}{1.0, 0.44, 0.37}
\definecolor{ballblue}{rgb}{0.13, 0.67, 0.8}
\definecolor{mydarkblue}{rgb}{0,0.08,0.45}
\definecolor{mydarkgreen}{RGB}{0, 139, 69}
\definecolor{MAEblue}{RGB}{47 112 182}
\definecolor{SDEblue}{RGB}{28 58 88}

\usepackage{adjustbox}
\definecolor{mycyan}{cmyk}{.3,0,0,0}
\usepackage{multirow}
\usepackage{array}
\usepackage{booktabs}
\usepackage{graphicx}
\usepackage{caption}

\title{\!\!Meta-{\color{orange}Un}learning on Diffusion Models: Preventing Relearning Unlearned Concepts\!\!\!}

\renewcommand\footnotemark{}
\author{
Hongcheng Gao$^{*1,2}$, Tianyu Pang$^{*\dagger1}$, Chao Du$^{1}$, Taihang Hu$^{4}$, Zhijie Deng$^{\dagger3}$, Min Lin$^{1}$ \thanks{\!\!$^*$Equal contribution. Work done during Hongcheng’s internship at Sea.}
\thanks{\!\!$^{\dagger}$Correspondence to Tianyu Pang and Zhijie Deng.}\\
  $^{1}$Sea AI Lab, Singapore $^{2}$University of Chinese Academy of Sciences \\
  $^{3}$Shanghai Jiao Tong University $^{4}$Nankai University \\
    \footnotesize{\texttt{\{gaohc, tianyupang, duchao, linmin\}@sea.com}; \texttt{zhijied@sjtu.edu.cn}}\\
}

\begin{document}

\maketitle
\begin{abstract}
With the rapid progress of diffusion models (DMs), significant efforts are being made to unlearn harmful or copyrighted concepts from pretrained DMs to prevent potential model misuse. However, it is observed that even when DMs are properly unlearned before release, malicious finetuning can compromise this process, causing DMs to \emph{relearn the unlearned concepts}. This occurs partly because certain benign concepts (e.g., ``skin'') retained in DMs are related to the unlearned ones (e.g., ``nudity''), facilitating their relearning via finetuning. To address this, we propose \textbf{meta-unlearning} on DMs. Intuitively, a meta-unlearned DM should behave like an unlearned DM when used as is; moreover, if the meta-unlearned DM undergoes malicious finetuning on unlearned concepts, the related benign concepts retained within it will be triggered to \emph{self-destruct}, hindering the relearning of unlearned concepts. Our meta-unlearning framework is compatible with most existing unlearning methods, requiring only the addition of an easy-to-implement meta objective. We validate our approach through empirical experiments on meta-unlearning concepts from Stable Diffusion models (SD-v1-4 and SDXL), supported by extensive ablation studies.

\end{abstract}

\vspace{-0.3cm}
\section{Introduction}
\vspace{-0.05cm}




Diffusion models (DMs) have achieved remarkable success in generative tasks~\citep{ho2020denoising,song2021score}, leading to the emergence of large-scale models like Stable Diffusion (SD) for text-to-image generation~\citep{rombach2022high}. However, training these models often requires vast datasets that may inadvertently contain private or copyrighted content, as well as harmful concepts that are not safe for work (NSFW)~\citep{schramowski2023safe}. These challenges have sparked interest in \emph{machine unlearning} algorithms for DMs~\citep{gandikota2023erasing,gandikota2024unified,kumari2023conceptablation,kim2023towards}, which modify pretrained models to forget specific inappropriate data (\emph{forget set}) while retaining performance on the remaining benign data (\emph{retain set}).

While unlearning methods designed for DMs are promising, recent studies reveal that unlearned models may be \textbf{maliciously finetuned} to \emph{relearn the unlearned concepts}, even when the finetuning is performed on unrelated benign data~\citep{qi2023fine,tamirisa2024tamper,patil2024can,shumailov2024ununlearning}. Although these studies focus primarily on language models, we observe similar phenomena on DMs as shown in Fig.~\ref{sdxl_nude}. This partly occurs because certain benign concepts (e.g., ``skin'' in the retain set) related to unlearned ones (e.g., ``nudity'' in the forget set) are still retained in DMs, easing their relearning during finetuning.

\begin{figure*}[t]
\vspace{-0.6cm}
\begin{center}
\includegraphics[width=0.84\linewidth]{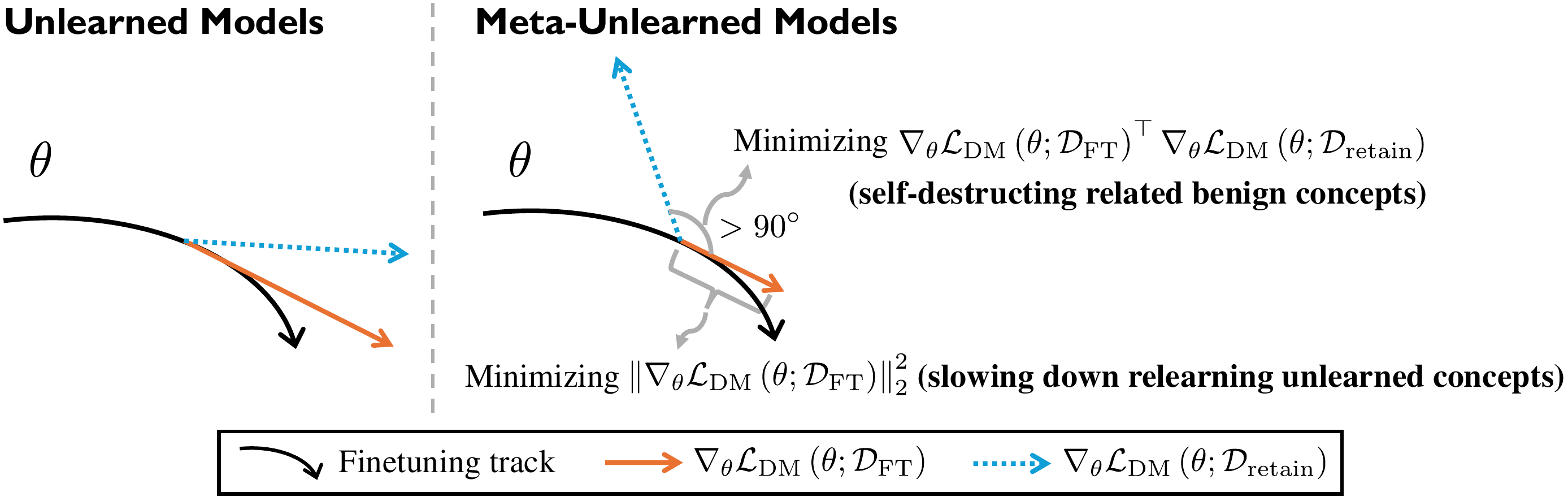}
\end{center}
\vspace{-0.5cm}
   \caption{{Mechanisms of finetuning} unlearned models (\emph{left}) and meta-unlearned models (\emph{right}) on a forget subset $\mathcal{D}_{\textrm{FT}}\subset\mathcal{D}_{\textrm{forget}}$. According to the first-order approximation described in Eq.~(\ref{equ99}), our meta-unlearning can slow down relearning unlearned concepts inside $\mathcal{D}_{\textrm{FT}}$, while self-destructing related benign concepts from $\mathcal{D}_{\textrm{retain}}$, i.e., $\mathcal{L}_{\textrm{DM}}(\theta;\mathcal{D}_{\textrm{retain}})$ increases when $\mathcal{L}_{\textrm{DM}}(\theta;\mathcal{D}_{\textrm{FT}})$ decreases.}
   \label{fig:demo}
   \vspace{-0.3cm}
\end{figure*}

To tackle this challenge, we draw inspiration from meta-learning~\citep{finn2017model} and propose the \textbf{meta-unlearning} framework. This framework comprises two key components: (1) a standard unlearning objective to ensure the model effectively forgets specified data before public release, while preserving performance on benign data; and (2) a \emph{meta objective} designed to slow down the relearning process if the model is maliciously finetuned on the forget set. Additionally, it induces the benign knowledge related to the forget set to self-destruct, as illustrated in Fig.~{\ref{fig:demo}}.

Our meta-unlearning framework is compatible with most existing unlearning methods for DMs, requiring only the addition of a simple-to-implement meta objective, as outlined in Algorithm~\ref{alg:metaunlearning}. This meta objective can be efficiently optimized by automatic differentiation~\citep{paszke2019pytorch}. We conduct extensive experiments on SD models (SD-v1-4 and SDXL) to validate the effectiveness of various instantiations of our meta-unlearning approach.



\vspace{-0.1cm}
\section{Related work}
\vspace{-0.05cm}

Recent studies have shown that DMs can be misused to generate unsafe content, such as images depicting sexual acts, harassment, or illegal activities~\citep{schramowski2023safe,gao2023evaluating,rando2022red}. To mitigate this issue, early-stage DMs are equipped with NSFW filters designed to block the generation of inappropriate images~\citep{rando2022red}. However, this approach does not prevent the model from generating harmful imagery at its core, and these filters can be bypassed, exposing security vulnerabilities~\citep{birhane2021multimodal,rombach2022high}.


\textbf{Machine unlearning on DMs.} Several methods have been proposed to unlearn or erase harmful, private, or copyrighted concepts from DMs~\citep{zhang2023forget,zhang2024defensive,park2024direct,huang2023receler,wu2024unlearning,pham2024robust,lu2024mace}. For instances, ESD~\citep{gandikota2023erasing} leverages negative guidance to finetune the U-Net, removing the specified style or concept. Concept ablation~\citep{kumari2023conceptablation} works by making the distribution of the target concept similar to that of an anchor concept. However, these methods are vulnerable to adversarial attacks. To this end, several adversarial-resistant unlearning methods have been proposed~\citep{li2024safegen,yang2024guardt2i,kim2024race,huang2024learning}. AdvUnlearn~\citep{zhang2024defensive} enhances the robustness of concept erasure by incorporating adversarial training principles, while RECE~\citep{gong2024reliable} derives new target embeddings for inappropriate content and iteratively aligns them with harmless concepts in cross-attention layers. Nevertheless, models unlearned by these methods can still be maliciously finetuned to relearn unlearned concepts, as observed in our experiments.


\textbf{Unlearned models can be attacked.} Recent studies have shown that unlearned models are vulnerable to generating previously unlearned concepts through adversarial attacks \citep{zhang2023generate,tsai2023ring,pham2023circumventing,ma2024jailbreaking, petsiuk2025concept, han2024probing, yuan2024towards} and malicious finetuning \citep{tamirisa2024tamper,shumailov2024ununlearning,Lucki2024adv,qi2023fine}. UnlearnDiffAtk~\citep{zhang2023generate} introduces an evaluation framework that uses adversarial attacks to generate adversarial prompts by exploiting the inherent classification capabilities of DMs. In the domain of language models, several works have revealed that finetuning can recover unlearned concepts. For example, \cite{qi2023fine} demonstrates that safety alignment and/or unlearning in language models can be undermined through finetuning with a small set of adversarially crafted training examples. Additionally, \cite{tamirisa2024tamper} show that refusal mechanisms and unlearning safeguards can be bypassed with minimal iterations of finetuning, while \cite{Lucki2024adv} recover most supposedly unlearned capabilities.\looseness=-1



\vspace{-0.2cm}
\section{Preliminaries}
\vspace{-0.2cm}
\label{sec3}
This section provides a brief review of diffusion models (DMs)~\citep{ho2020denoising,song2021score} and commonly used machine unlearning methods in the DM literature.

\subsection{Diffusion Models}
\label{sec31}
We focus on discrete-time DMs that serve as the cornerstone of Stable Diffusion~\citep{rombach2022high}.
We consider random variables $\boldsymbol{x}\in\mathcal{X}$ and $\boldsymbol{c}\in\mathcal{C}$, where $\boldsymbol{x}$ denotes the latent feature and $\boldsymbol{c}$ the conditional context, e.g., text prompts. 
Let $q(\boldsymbol{x},\boldsymbol{c})$ denote the data distribution. 
Consider a \emph{forward} diffusion process over time interval $[0, T]$ with $T\in \mathbb{N}^{+}$.
The Markov transition probability from $\boldsymbol{x}_{t-1}$ to $\boldsymbol{x}_{t}$ is $q(\boldsymbol{x}_{t}|\boldsymbol{x}_{t-1})\triangleq\mathcal{N}(\boldsymbol{x}_{t}|\sqrt{1-\beta_{t}}\boldsymbol{x}_{t-1},\beta_{t}\mathbf{I})$, where $\boldsymbol{x}_{0}=\boldsymbol{x}$ and $\beta_{1},\cdots,\beta_{T}$ correspond to a variance schedule. Note that we can sample $\boldsymbol{x}_{t}$ at an arbitrary timestep $t$ directly from $\boldsymbol{x}$, since there is $q(\boldsymbol{x}_{t}|\boldsymbol{x})=\mathcal{N}(\boldsymbol{x}_{t}|\sqrt{\overline{\alpha}_{t}}\boldsymbol{x},(1-\overline{\alpha}_{t})\mathbf{I})$,
where $\alpha_{t}\triangleq 1-\beta_{t}$ and $\overline{\alpha}_{t}\triangleq \prod_{i=1}^{t}\alpha_{i}$.

\citet{sohl2015deep} show that when $\beta_{t}$ are small, the \emph{reverse} diffusion process can also be modeled by Gaussian conditionals. Specifically, the reverse transition probability from $\boldsymbol{x}_{t}$ to $\boldsymbol{x}_{t-1}$ is written as $p_{\theta}(\boldsymbol{x}_{t-1}|\boldsymbol{x}_{t},\boldsymbol{c})=\mathcal{N}(\boldsymbol{x}_{t-1}|\boldsymbol{\mu}_{\theta}(\boldsymbol{x}_{t},\boldsymbol{c}),\sigma_{t}^{2}\mathbf{I})$, where $\theta\in\mathbb{R}^{d}$ is the model parameters and $\sigma_{t}$ are time dependent constants. Instead of directly modeling the data prediction $\boldsymbol{\mu}_{\theta}$, we choose to model the noise prediction $\boldsymbol{\epsilon}_{\theta}$ based on the parameterization $\boldsymbol{\mu}_{\theta}(\boldsymbol{x}_{t},\boldsymbol{c})=\frac{1}{\sqrt{\alpha_{t}}}\left(\boldsymbol{x}_{t}-\frac{\beta_{t}}{\sqrt{1-\overline{\alpha}_{t}}}\boldsymbol{\epsilon}_{\theta}(\boldsymbol{x}_{t},\boldsymbol{c})\right)$.
The training objective of $\boldsymbol{\epsilon}_{\theta}(\boldsymbol{x}_{t},\boldsymbol{c})$ can be derived from optimizing the (weighted) variational bound of negative log-likelihood, formulated as follows:
\begin{equation*}
    \min_{\theta}\mathcal{L}_{\textrm{DM}}(\theta;\mathcal{D}_{\textrm{train}})=\mathbb{E}_{(\boldsymbol{x},\boldsymbol{c})\sim\mathcal{D}_{\textrm{train}},\boldsymbol{\epsilon},t}\left[\left\|\boldsymbol{\epsilon}-\boldsymbol{\epsilon}_{\theta}(\boldsymbol{x}_{t},\boldsymbol{c})\right\|_{2}^{2}\right]\textrm{,}
\end{equation*}
where $\boldsymbol{x}_{t}=\sqrt{\overline{\alpha}_{t}}\boldsymbol{x}+\sqrt{1-\overline{\alpha}_{t}}\boldsymbol{\epsilon}$, the pairs $(\boldsymbol{x},\boldsymbol{c})$ are sampled from the training set $\mathcal{D}_{\textrm{train}}$, $\boldsymbol{\epsilon}\sim \mathcal{N}(\boldsymbol{\epsilon}|\mathbf{0},\mathbf{I})$ is a standard Gaussian noise, and $t\sim \mathcal{U}([1,T])$ is a uniform distribution.

\subsection{Machine Unlearning for DMs}
\label{sec32}
DMs, despite their high capability, may generate unsafe content or disclose sensitive information that is not safe for work (NSFW)~\citep{schramowski2023safe}. Several recent studies have investigated concept erasing or machine unlearning for DMs to address safety, privacy, and copyright concerns~\citep{kumari2023conceptablation,zhang2024defensive,heng2024selective}. Let $\boldsymbol{\epsilon}_{\theta^{*}}$ denotes the DM pretrained on the dataset $\mathcal{D}_{\textrm{train}}$, where $\theta^*=\arg\min_{\theta}\mathcal{L}_{\textrm{DM}}(\theta;\mathcal{D}_{\textrm{train}})$. The goal of machine unlearning is to unlearn a \emph{forget} set $\mathcal{D}_{\text{forget}}\subset\mathcal{D}_{\textrm{train}}$ from $\boldsymbol{\epsilon}_{\theta^{*}}$, while preserving performance on the \emph{retain} set $\mathcal{D}_{\text{retain}}=\mathcal{D}_{\text{train}}\backslash\mathcal{D}_{\text{forget}}$. Below we describe four unlearning methods for DMs that we use as baselines.

\textbf{Erased Stable Diffusion (ESD)}~\citep{gandikota2023erasing} intervenes pretrained DMs by steering generation away from the concept intended to be forgotten. Ideally, the unlearned DM is expected to predict $\widetilde{\boldsymbol{\epsilon}}_{\theta^{*}}(\boldsymbol{x}_{t},\boldsymbol{c})=\boldsymbol{\epsilon}_{\theta^{*}}(\boldsymbol{x}_{t},\emptyset)-\eta\left[\boldsymbol{\epsilon}_{\theta^{*}}(\boldsymbol{x}_{t},\boldsymbol{c})-\boldsymbol{\epsilon}_{\theta^{*}}(\boldsymbol{x}_{t},\emptyset)\right]$ when fed in $(\boldsymbol{x},\boldsymbol{c})\sim\mathcal{D}_{\textrm{forget}}$, where $\eta>0$ is a hyperparameter and $\emptyset$ indicates unconditional context. The unlearning objective of ESD is
    \begin{equation*}
        \begin{split}
            &\min_{\theta}\mathcal{L}_{\textrm{ESD}}(\theta;\mathcal{D}_{\textrm{forget}})\\
            =&\mathbb{E}_{(\boldsymbol{x},\boldsymbol{c})\sim\mathcal{D}_{\textrm{forget}},\boldsymbol{\epsilon},t}\left[\left\|\boldsymbol{\epsilon}_{\theta}(\boldsymbol{x}_{t},\boldsymbol{c})-\widetilde{\boldsymbol{\epsilon}}_{\theta^{*}}(\boldsymbol{x}_{t},\boldsymbol{c})\right\|_{2}^{2}\right]\textrm{,}
        \end{split}
    \end{equation*}
    where $\theta$ is initialized from the frozen $\theta^{*}$. \citet{gandikota2023erasing} use ESD-x-$\eta$ to indicate only cross-attention parameters are finetuned with hyperparameter $\eta$; likewise, ESD-u-$\eta$ indicates only non-cross-attention parameters are finetuned, and ESD-f-$\eta$ indicates full finetuning.

\textbf{Safe self-distillation diffusion (SDD)}~\citep{kim2023towards} is a self-distillation paradigm to erase concepts from DMs. The unlearning objective of SDD is to optimize
    \begin{equation*}
        \begin{split}
            &\min_{\theta}\mathcal{L}_{\textrm{SDD}}(\theta;\mathcal{D}_{\textrm{forget}})\\
            =&\mathbb{E}_{(\boldsymbol{x},\boldsymbol{c})\sim\mathcal{D}_{\textrm{forget}},\boldsymbol{\epsilon},t}\left[\left\|\boldsymbol{\epsilon}_{\theta}(\boldsymbol{x}_{t},\boldsymbol{c})-\texttt{sg}\left(\boldsymbol{\epsilon}_{\theta}(\boldsymbol{x}_{t},\emptyset)\right)\right\|_{2}^{2}\right]\textrm{,}
        \end{split}
    \end{equation*}
    where $\texttt{sg}$ is the  stop-gradient operation and $\theta$ is initialized from the frozen $\theta^{*}$. To mitigate catastrophic forgetting, SDD employs an EMA teacher. Note that in the original implementations of both ESD and SDD, there are only text prompts $\boldsymbol{c}$ in the forget set, while the noisy latents $\boldsymbol{x}_{t}$ are generated by the frozen DM $\boldsymbol{\epsilon}_{\theta^*}$.

    \textbf{Unified concept editing (UCE)}~\citep{gandikota2024unified} edits the pretrained DMs via a closed-form solution without finetuning. Let $W^{*}$ be the attention matrices of $\theta^{*}$ (Key/Value matrices), $\mathcal{T}$ be the text embedding mapping in $\boldsymbol{\epsilon}_{\theta^{*}}$, then the unlearning objective of UCE is to optimize
    \begin{equation*}
    \begin{split}
        \min_{W}&\mathbb{E}_{\boldsymbol{c}_{f},\boldsymbol{c}_{r}}\Big[\left\|W\mathcal{T}(\boldsymbol{c}_{f})-W^{*}\mathcal{T}(\emptyset)\right\|_{2}^{2}+\\
    &\lambda_{1}\left\|W\mathcal{T}(\boldsymbol{c}_{r})-W^{*}\mathcal{T}(\boldsymbol{c}_{r})\right\|_{2}^{2}+\lambda_{2}\left\|W-W^{*}\right\|_{2}^{2}\Big]\textrm{,}
    \end{split}
    \end{equation*}
    where $\boldsymbol{c}_{f}\sim\mathcal{D}_{\textrm{forget}}$, $\boldsymbol{c}_{r}\sim\mathcal{D}_{\textrm{retain}}$, and $\lambda_{1}$, $\lambda_{2}$ are hyperparameters. \citet{gandikota2024unified} prove that the above minimization problem has closed-form solution $W_{\textrm{UCE}}$.

\textbf{Reliable and efficient concept erasure (RECE)}~\citep{gong2024reliable} first performs UCE, after which iteratively creates new erasing embeddings and obtains updated attention matrices. Specifically, let $\widetilde{W}\leftarrow W_{\textrm{UCE}}$, and use subscripts $i$ to denote the $i$-th attention matrix in the model; then RECE iteratively constructs $\boldsymbol{c}'$ by optimizing 
    \begin{equation*}
        \min_{\boldsymbol{c}'}\sum_{i}\left\|\widetilde{W}_{i}\mathcal{T}(\boldsymbol{c}')-W_{i}^{*}\mathcal{T}(\boldsymbol{c}_{f})\right\|_{2}^{2}+\lambda\left\|\mathcal{T}(\boldsymbol{c}')\right\|_{2}^{2}\textrm{,}
    \end{equation*}
    where $\lambda$ is a hyperparameter. The constructed $\boldsymbol{c}'$ is used to derive $\widetilde{W}'$ by UCE, then update as $\widetilde{W}\leftarrow \widetilde{W}'$ and finally obtain $W_{\textrm{RECE}}=\widetilde{W}$.

\begin{algorithm}[t]
\caption{The general framework of \emph{meta-unlearning}}\label{alg:metaunlearning}
\begin{algorithmic}[1]
\Require Pretrained $\theta^{*}$, forget set $\mathcal{D}_{\text{forget}}$, retain set $\mathcal{D}_{\text{retain}}$
\Require Unlearning objective $\mathcal{L}_{\textrm{unlearn}}$, finetuning objective $\mathcal{L}_{\textrm{FT}}$, meta objective $\mathcal{L}_{\textrm{meta}}$
\Require Outer (steps $N$, learning rate $\omega$), inner (steps $M$, learning rate $\tau$), scale factors $\gamma_{1}, \gamma_{2}$
\item[] {\# {\color{darkpowderblue}If $\mathcal{L}_{\textrm{unlearn}}$ is ESD/SDD that needs optimization}}
\State {\color{darkpowderblue}$\theta_0 \gets \theta^*$}
\item[] {\# {\color{ballblue}If $\mathcal{L}_{\textrm{unlearn}}$ is UCE/RECE that has closed-form solution}}
\State {\color{ballblue}$\theta_0 \gets \theta^{\textrm{UN}}=\argmin_{\theta}\mathcal{L}_{\textrm{unlearn}}$}
\For{$n = 1$ to $N$}
    \State Sample a finetuning set $\mathcal{D}_{\text{FT}}\subset\mathcal{D}_{\text{forget}}$
    \State Initialize $\bm{g}=\bm{0}$ and $\theta^{\textrm{FT}}=\theta_{n-1}$
    \item[] \hspace{0.45cm} {\# {\color{darkpowderblue}If $\mathcal{L}_{\textrm{unlearn}}$ is ESD/SDD}}
    \State {\color{darkpowderblue}$\bm{g}\gets\bm{g}+\gamma_{1}\cdot\nabla_{\theta_{n-1}} \mathcal{L}_{\text{unlearn}}(\theta_{n-1}; \mathcal{D}_{\text{forget}},\mathcal{D}_{\text{retain}})$}
    \For{$m = 1$ to $M$}
    \State $\theta^{\textrm{FT}}\gets\theta^{\textrm{FT}}-\tau\cdot\nabla_{\theta^{\textrm{FT}}}\mathcal{L}_{\textrm{FT}}\left(\theta;\mathcal{D}_{\textrm{FT}}\right)$ 
    \EndFor
    \item[] \hspace{0.45cm} {\# {\color{bittersweet}Meta objective}} 
    \State {\color{bittersweet}$\bm{g} \gets \bm{g}+\gamma_{2}\cdot\nabla_{\theta_{n-1}} \mathcal{L}_{\text{meta}}(\theta^{\textrm{FT}}; \mathcal{D}_{\text{FT}},\mathcal{D}_{\text{retain}})$}
    \State $\theta_n \gets \theta_{n-1} - \omega\cdot \bm{g}$
\EndFor
\State \Return $\theta_N$
\end{algorithmic}
\end{algorithm}

\begin{figure*}[t]
\begin{center}
\vspace{-0.5cm}
\includegraphics[width=.82\linewidth]{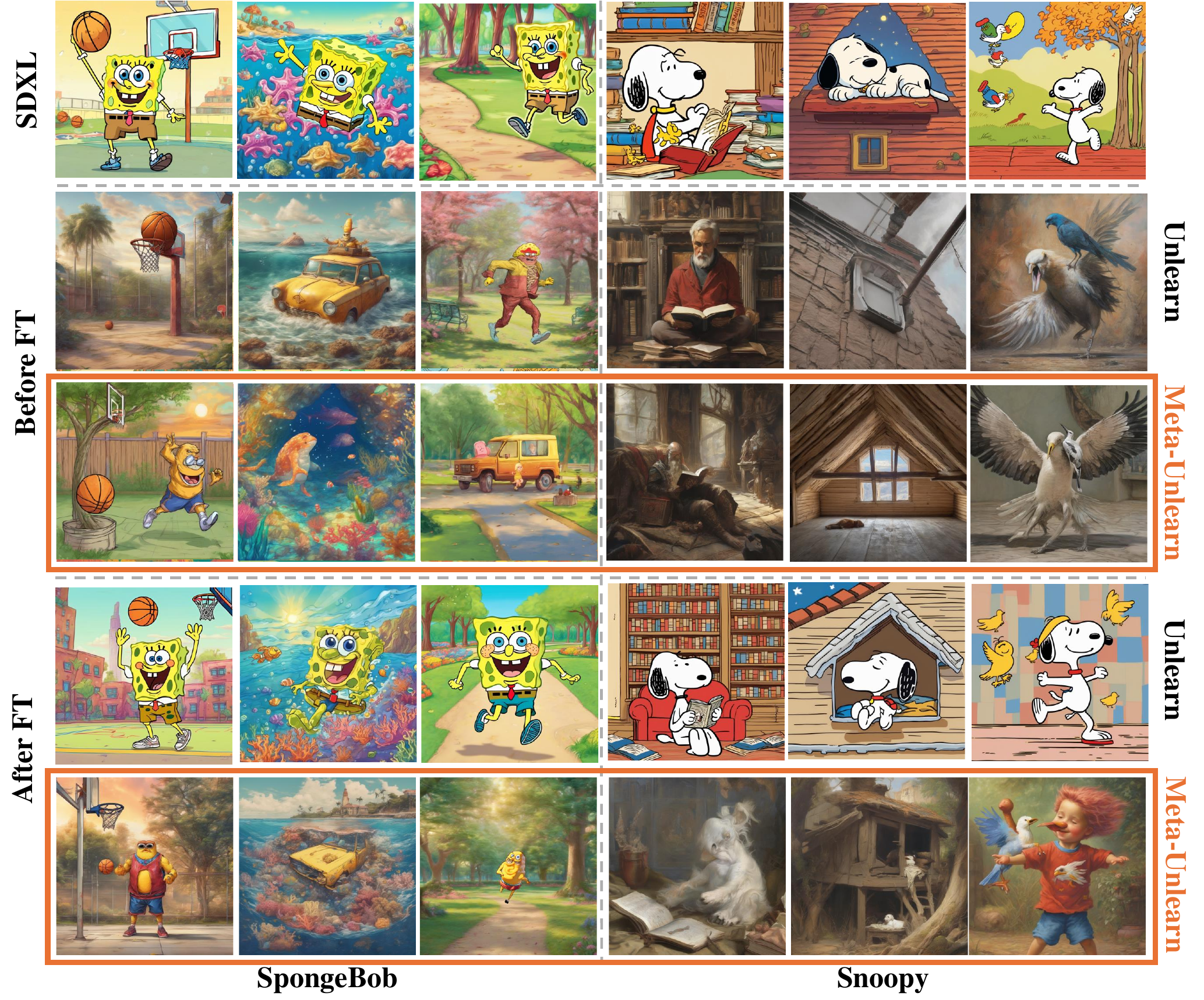}
\end{center}
\vspace{-0.6cm}
   \caption{\textbf{Images generated by copyright prompts.} The top panel displays images generated using the original SDXL model for copyright prompts. In the following panels, we show images generated using ESD-u-1 \emph{unlearned} and \emph{meta-unlearned} SDXL models before and after finetuning (FT) on the dataset of unlearned copyright concept for 100 steps, respectively. The forget copyright concept for the left three columns is ``\texttt{SpongeBob}'', while the right is ``\texttt{Snoopy}''.} 
   \label{sdxl_copyright}
   \vspace{-0.1cm}
\end{figure*}

\vspace{-0.1cm}
\section{Meta-unlearning for DMs}
\vspace{-0.1cm}
\label{meta_framework}

In Section~\ref{sec32}, we briefly introduce the commonly used unlearning methods for DMs. 
In general, the unlearning objective can be summarized as forgetting knowledge from $\mathcal{D}_{\textrm{forget}}$ and preserving performance on $\mathcal{D}_{\textrm{retain}}$, i.e., solving 
\vspace{-0.2cm}
\begin{equation}
\begin{split}
&\min_{\theta}\mathcal{L}_{\textrm{unlearn}}\left(\theta;\mathcal{D}_{\textrm{forget}},\mathcal{D}_{\textrm{retain}}\right)\\
\triangleq &\mathcal{L}_{\textrm{forget}}\left(\theta;\mathcal{D}_{\textrm{forget}}\right)+\lambda\cdot\mathcal{L}_{\textrm{retain}}\left(\theta;\mathcal{D}_{\textrm{retain}}\right)\textrm{,}
\end{split}
    \label{equ66}
\end{equation}
where $\mathcal{L}_{\textrm{forget}}$ is to unlearn the forget set, $\mathcal{L}_{\textrm{retain}}$ is to keep performance on the retain set, and $\lambda$ is a trade-off hyperparameter. Various unlearning methods correspond to different instantiations of $\mathcal{L}_{\textrm{forget}}$ and $\mathcal{L}_{\textrm{retain}}$. In particular, ESD and SDD require optimizers to solve Eq.~(\ref{equ66}), whereas UCE and RECE have closed-form solutions. In optimization, the initialized parameters are usually set to the pretrained $\theta^{*}$ and the unlearned parameters are denoted as $\theta^{\textrm{UN}}$.

\vspace{-0.1cm}
\subsection{Meta-unlearning framework}
\vspace{-0.05cm}
A publicly released DM can potentially be finetuned to adapt to various downstream tasks. However, as observed in previous studies, finetuning or modifying weights of a model could comprise its alignment and/or unlearning~\citep{qi2023fine,tamirisa2024tamper}. This underscores the need for mechanisms to \emph{simulate the finetuning process in advance}, ensuring DMs are resilient against relearning the unlearned concepts. Inspired by meta-learning~\citep{finn2017model}, we propose the \textbf{meta-unlearning} framework, as illustrated in Algorithm~\ref{alg:metaunlearning}. Our framework consists of two components: (1) the standard unlearning objective $\mathcal{L}_{\textrm{unlearn}}$, as described above, and (2) the meta objective $\mathcal{L}_{\textrm{meta}}$, which \emph{resists the relearning of unlearned concepts}, even after finetuning on the forget set.

Formally, we define $\mathcal{L}_{\textrm{FT}}$ as the finetuning objective, and let $\mathcal{D}_{\textrm{FT}}\subset\mathcal{D}_{\textrm{forget}}$ represent the \emph{malicious finetuning} dataset, which is designed to intentionally make the model relearn concepts from the forget set. The finetuned model parameters $\theta^{\textrm{FT}}$ are updated by one or more gradient descents. For example, when using one gradient update from $\theta$, there is $ \theta^{\textrm{FT}}\leftarrow\theta-\tau\cdot\nabla_{\theta}\mathcal{L}_{\textrm{FT}}\left(\theta;\mathcal{D}_{\textrm{FT}}\right)$, where $\tau$ is the step size. The parameters $\theta$ is trained by the meta objective $\mathcal{L}_{\textrm{meta}}$ as:
\vspace{-0.1cm}
\begin{equation}\label{equ77}
    \begin{split}
        &\min_{\theta} \mathcal{L}_{\textrm{meta}}(\theta^{\textrm{FT}}; \mathcal{D}_{\text{FT}},\mathcal{D}_{\text{retain}})\\
        =&\mathcal{L}_{\textrm{meta}}(\theta-\tau\cdot\nabla_{\theta}\mathcal{L}_{\textrm{FT}}\left(\theta;\mathcal{D}_{\textrm{FT}}\right); \mathcal{D}_{\text{FT}},\mathcal{D}_{\text{retain}})\textrm{.}
    \end{split}
\end{equation}
To optimize the RHS of Eq.~(\ref{equ77}), the gradients are back-propagated through both $\theta$ and $\nabla_{\theta}\mathcal{L}_{\textrm{FT}}\left(\theta;\mathcal{D}_{\textrm{FT}}\right)$ that can be efficiently computed by automatic differentiation~\citep{paszke2019pytorch}.

\begin{figure*}[t]
\begin{center}
\vspace{-0.5cm}
\includegraphics[width=.82\linewidth]{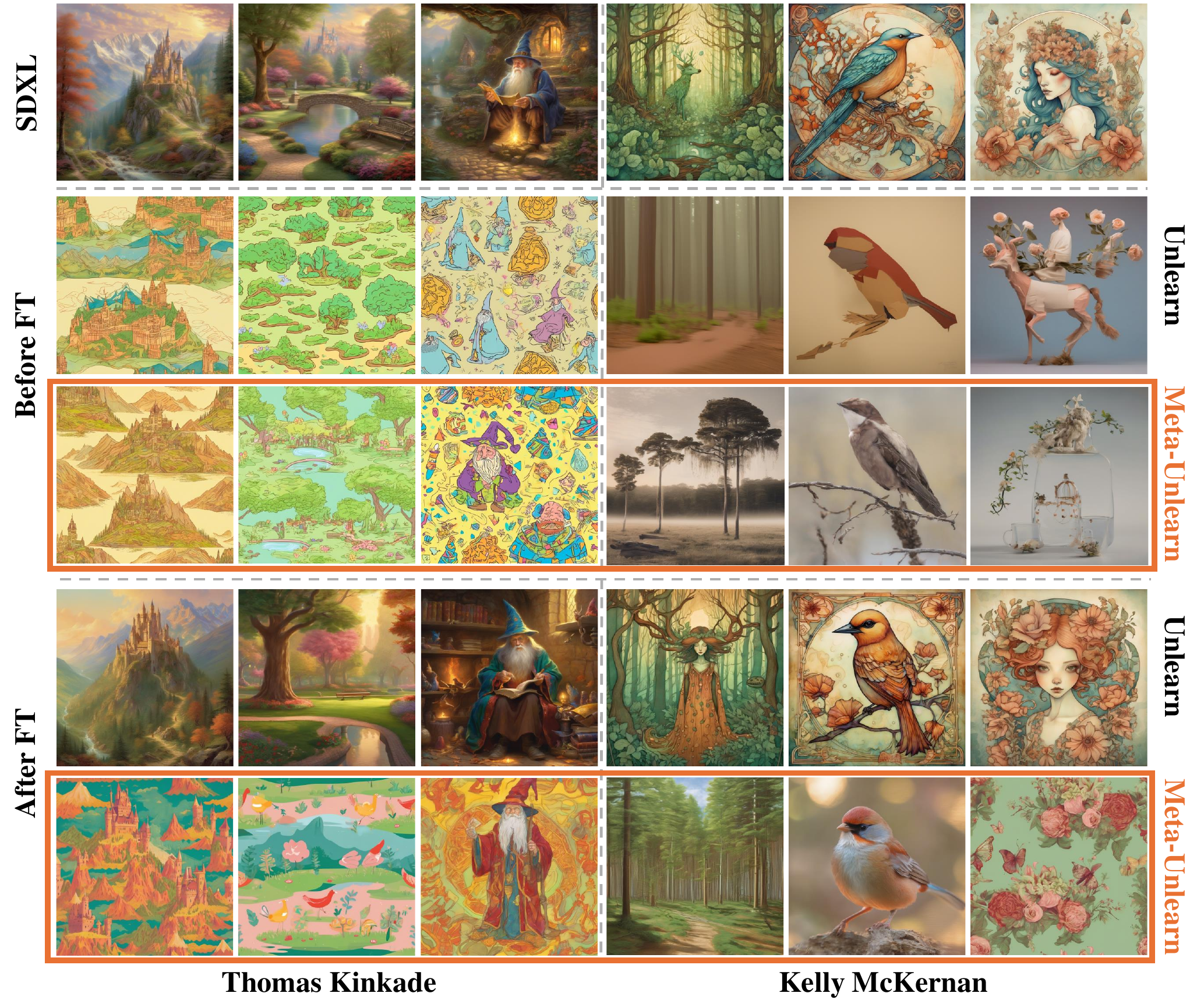}
\end{center}
\vspace{-0.6cm}
   \caption{\textbf{Images generated by style prompts.} The top panel displays images generated using the original SDXL model for style prompts. In the following panels, we show images generated using ESD-u-1 \emph{unlearned} and \emph{meta-unlearned} SDXL models before and after finetuning (FT) on the dataset of unlearned copyright concept for 100 steps, respectively. The forget style of left three columns is ``\texttt{Thomas Kinkade}'' and the right is ``\texttt{Kelly McKernan}''.} 
   \label{sdxl_style}
   \vspace{-0.1cm}
\end{figure*}

\vspace{-0.05cm}
\subsection{Meta objective \texorpdfstring{$\mathcal{L}_{\textrm{meta}}$}{TEXT}}
\vspace{-0.05cm}
Our design goal for meta-unlearning is to ensure that after the model is maliciously finetuned on $\mathcal{D}_{\textrm{FT}}\subset\mathcal{D}_{\textrm{forget}}$, it cannot relearn the unlearned concepts. Additionally, we further encourage the model to \emph{self-destruct knowledge from the retain set}. Given this, a natural instantiation of $\mathcal{L}_{\textrm{meta}}$ is
\begin{equation}\label{eq8}
    \!\!\!\!\!\!\!\!\begin{split}
        \min_{\theta}\mathcal{L}_{\textrm{meta}}&(\theta^{\textrm{FT}}; \mathcal{D}_{\text{FT}},\mathcal{D}_{\text{retain}})\triangleq -\mathcal{L}_{\textrm{DM}}(\theta^{\textrm{FT}};\mathcal{D}_{\textrm{FT}})\\
        &-\zeta\cdot\left[\mathcal{L}_{\textrm{DM}}(\theta^{\textrm{FT}};\mathcal{D}_{\textrm{retain}})-\mathcal{L}_{\textrm{DM}}(\theta;\mathcal{D}_{\textrm{retain}})\right]\textrm{,}\!\!\!\!
    \end{split}
\end{equation}
where $\mathcal{L}_{\textrm{DM}}$ is the diffusion loss described in Section~\ref{sec31} and $\zeta$ is a hyperparameter. Now we take a close look at how the meta objective in Eq.~(\ref{eq8}) works. In practice, the finetuning objective $\mathcal{L}_{\textrm{FT}}$ is typically set to $\mathcal{L}_{\textrm{DM}}$; and following Eq.~(\ref{equ77}), the first-order approximation of $\mathcal{L}_{\textrm{meta}}(\theta^{\textrm{FT}}; \mathcal{D}_{\text{FT}}, \mathcal{D}_{\text{retain}})$ can be written as (up to a $\mathcal{O}(\tau^{2})$ error)
\begin{equation}
   \begin{split}
        \mathcal{L}_{\textrm{meta}}\approx&-\mathcal{L}_{\textrm{DM}}(\theta;\mathcal{D}_{\textrm{FT}})+\tau\cdot{\color{bittersweet}\left\|\nabla_{\theta}\mathcal{L}_{\textrm{DM}}\left(\theta;\mathcal{D}_{\textrm{FT}}\right)\right\|_{2}^{2}}\\
        &+\tau\zeta\cdot{\color{bittersweet}\nabla_{\theta}\mathcal{L}_{\textrm{DM}}\left(\theta;\mathcal{D}_{\textrm{FT}}\right)^{\top}\nabla_{\theta}\mathcal{L}_{\textrm{DM}}\left(\theta;\mathcal{D}_{\textrm{retain}}\right)}\textrm{,}
   \end{split}
   \label{equ99}
\end{equation}
where we colorize the terms that play key roles in our meta-unlearning framework. Note that this approximation corresponds to $M=1$ in Algorithm~\ref{alg:metaunlearning}; for $M>1$ (i.e., multi-step gradient descent), the approximation formula remains unchanged but with equivalent step size $M\tau$.

\begin{figure*}[t]
\begin{center}
\vspace{-0.5cm}
\includegraphics[width=.770\linewidth]{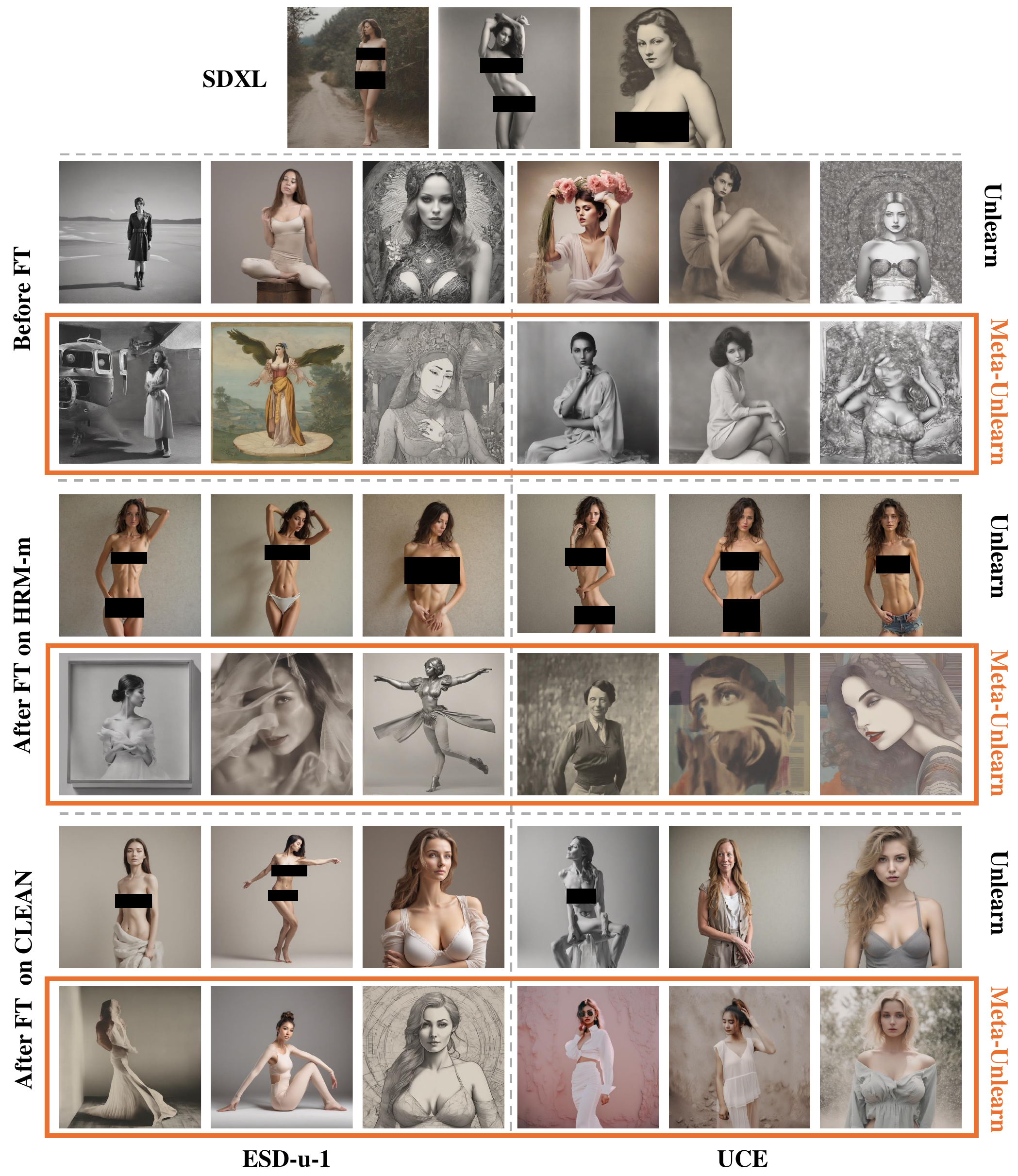}
\end{center}
\vspace{-0.65cm}
   \caption{\textbf{Images generated by harmful prompts.} The top panel displays images generated using the original SDXL model for harmful prompts. Following panels show the images generated using \emph{unlearned} and \emph{meta-unlearned} SDXL models before finetuning (FT), after FT on the HRM-m dataset for 100 steps, and after FT on the CLEAN dataset for 100 steps, respectively. The left three columns display images generated by ESD-u-1 and its meta variant, while the right three columns display images generated by UCE and its meta variant.} 
   \label{sdxl_nude}
   \vspace{-0.4cm}
\end{figure*}

\textbf{Remark.} As in Fig.~\ref{fig:demo}, minimizing $\left\|\nabla_{\theta}\mathcal{L}_{\textrm{DM}}\left(\theta;\mathcal{D}_{\textrm{FT}}\right)\right\|_{2}^{2}$ decreases gradient norm and thus delays relearning of forget set. Minimizing $\nabla_{\theta}\mathcal{L}_{\textrm{DM}}\left(\theta;\mathcal{D}_{\textrm{FT}}\right)^{\top}\nabla_{\theta}\mathcal{L}_{\textrm{DM}}\left(\theta;\mathcal{D}_{\textrm{retain}}\right)$ induces a $>90^\circ$ angle between the gradients $\nabla_{\theta}\mathcal{L}_{\textrm{DM}}\left(\theta;\mathcal{D}_{\textrm{FT}}\right)$ and $\nabla_{\theta}\mathcal{L}_{\textrm{DM}}\left(\theta;\mathcal{D}_{\textrm{retain}}\right)$, such that when the model is finetuned along $\nabla_{\theta}\mathcal{L}_{\textrm{DM}}\left(\theta;\mathcal{D}_{\textrm{FT}}\right)$ (the loss $\mathcal{L}_{\textrm{DM}}\left(\theta;\mathcal{D}_{\textrm{FT}}\right)$ decreases), the knowledge inside the retain set will self-destruct (namely, the loss $\mathcal{L}_{\textrm{DM}}\left(\theta;\mathcal{D}_{\textrm{retain}}\right)$ increases).

\begin{table*}[t]
\vspace{-0.45cm}
\centering
\caption{\textbf{Nudity evaluation.} The nudity score of \emph{unlearned} and \emph{meta-unlearned} SD-v1-4 models, based on six unlearning methods similar to Table~\ref{before}. The results are reported for models before or after finetuning (FT) on three datasets for 50, 100, 200, and 300 steps.}
\vspace{-0.2cm}
\renewcommand*{\arraystretch}{1.2}
\setlength{\tabcolsep}{2.6pt}
\adjustbox{max width=0.9\textwidth}{
\begin{tabular}{cc|c|cccc|cccc|cccc}
\toprule
\multicolumn{1}{l}{}                                        &                                 & {\color[HTML]{1F2328} {\ul \textbf{Before FT}}} & \multicolumn{4}{c|}{{\color[HTML]{1F2328} {\ul \textbf{FT on HRM-m}}}} & \multicolumn{4}{c|}{{\color[HTML]{1F2328} {\ul \textbf{FT on HRM-s}}}} & \multicolumn{4}{c}{{\color[HTML]{1F2328} {\ul \textbf{FT on CLEAN}}}} \\
\multicolumn{1}{l}{\multirow{-2}{*}{\textbf{Method}}} & \multirow{-2}{*}{\textbf{Type}} & 0                                               & 50               & 100             & 200             & 300             & 50               & 100             & 200             & 300             & 50              & 100             & 200             & 300             \\
\midrule
SD-1.4                                                      & -                               & 97.18                                           & -                & -               & -               & -               & -                & -               & -               & -               & -               & -               & -               & -               \\
\rowcolor[HTML]{D9D9D9} 
\cellcolor[HTML]{D9D9D9}                                    & Unlearn                         & 6.34                                            & 19.01            & 21.83           & 30.28           & 34.51           & 23.24            & 24.65           & 45.07           & 53.52           & 11.27           & 13.38           & 12.68           & 14.79           \\
\rowcolor[HTML]{D9D9D9} 
\multirow{-2}{*}{\cellcolor[HTML]{D9D9D9}ESD-u-1}           & Meta-Unlearn                           & 0.00                                            & 8.45             & 13.38           & 23.94           & 26.06           & 4.23             & 12.68           & 30.28           & 38.03           & 2.82            & 2.11            & 4.23            & 4.23            \\
                                                            & Unlearn                         & 3.52                                            & 26.76            & 38.73           & 36.62           & 33.80           & 23.24            & 28.17           & 31.69           & 35.92           & 5.63            & 4.93            & 7.75            & 6.34            \\
\multirow{-2}{*}{ESD-u-3}                                   & Meta-Unlearn                            & 0.00                                            & 3.52             & 19.01           & 26.76           & 26.76           & 8.45             & 18.31           & 20.42           & 26.06           & 2.11            & 2.82            & 4.23            & 2.82            \\
\rowcolor[HTML]{D9D9D9} 
\cellcolor[HTML]{D9D9D9}                                    & Unlearn                         & 6.34                                            & 32.39            & 56.34           & 60.56           & 55.63           & 47.89            & 51.41           & 40.14           & 59.86           & 12.68           & 16.90           & 18.31           & 14.79           \\
\rowcolor[HTML]{D9D9D9} 
\multirow{-2}{*}{\cellcolor[HTML]{D9D9D9}ESD-f-3}           & Meta-Unlearn                            & 0.00                                            & 2.11             & 26.06           & 38.03           & 33.10           & 5.63             & 18.31           & 24.65           & 35.92           & 3.52            & 4.93            & 5.63            & 5.63            \\
                                                            & Unlearn                         & 1.41                                            & 33.10            & 57.04           & 52.11           & 54.23           & 42.96            & 50.70           & 53.52           & 53.52           & 14.08           & 16.20           & 17.61           & 18.31           \\
\multirow{-2}{*}{SDD}                                       & Meta-Unlearn                            & 0.00                                            & 20.42            & 45.07           & 42.25           & 48.59           & 15.49            & 28.17           & 31.69           & 35.21           & 2.11            & 5.63            & 6.34            & 7.75            \\
\rowcolor[HTML]{D9D9D9} 
\cellcolor[HTML]{D9D9D9}                                    &Unlearn                           & 16.90                                           & 36.62            & 44.37           & 47.89           & 36.62           & 28.17            & 34.51           & 40.14           & 57.75           & 23.94           & 25.35           & 23.24           & 26.76           \\
\rowcolor[HTML]{D9D9D9} 
\multirow{-2}{*}{\cellcolor[HTML]{D9D9D9}UCE}               & Meta-Unlearn                            & 1.41                                            & 24.65            & 28.17           & 30.28           & 25.35           & 18.31            & 19.01           & 21.13           & 42.96           & 4.93            & 5.63            & 4.93            & 4.23            \\
                                                            & Unlearn                         & 4.93                                            & 16.20            & 19.72           & 22.54           & 22.54           & 11.27            & 14.79           & 17.61           & 22.54           & 6.34            & 9.86            & 7.04            & 7.75            \\
\multirow{-2}{*}{RECE}                                      & Meta-Unlearn                            & 4.23                                            & 7.04             & 10.56           & 15.49           & 13.38           & 5.63             & 8.45            & 13.38           & 15.49           & 4.23            & 5.63            & 4.93            & 5.63            \\ \bottomrule
\end{tabular}\label{after}}
\vspace{-0.3cm}
\end{table*}

\begin{figure}[t]
\begin{center}
\includegraphics[width=1.\linewidth]{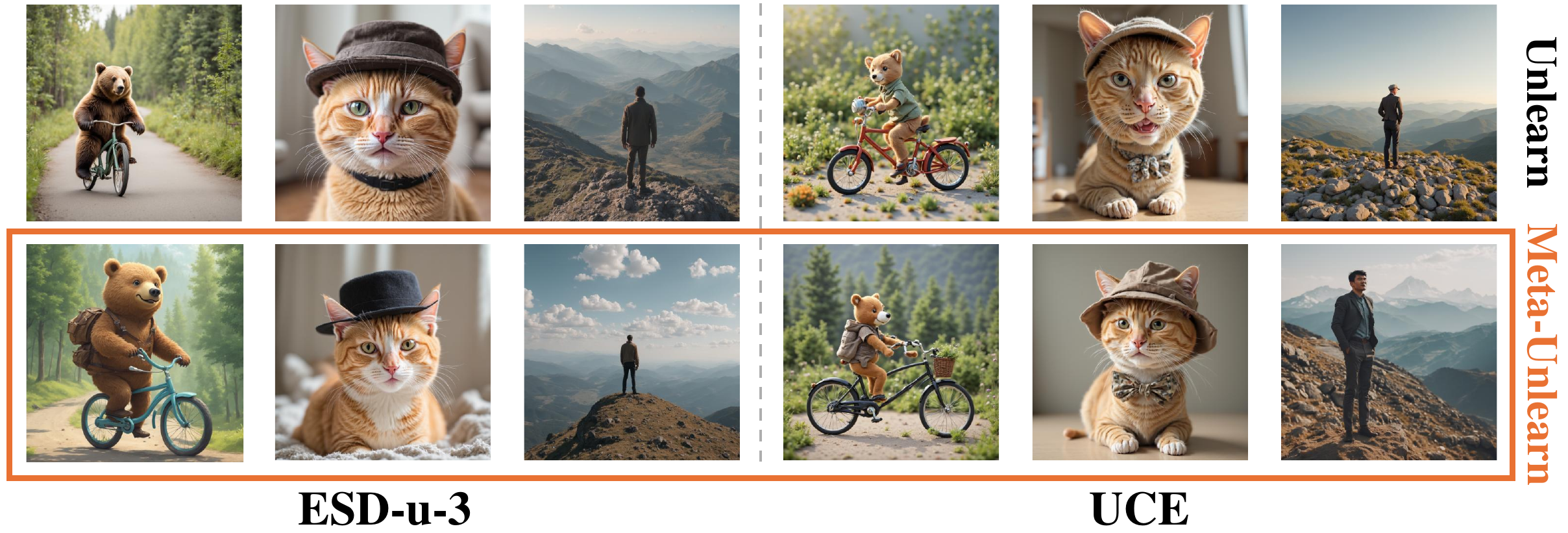}
\end{center}
\vspace{-0.6cm}
   \caption{\textbf{Images generated by benign prompts after finetuning on CLEAN.} \emph{Unlearned} and \emph{meta-unlearned} SDXL models are finetuned on the CLEAN dataset for 100 steps. As seen, our methods will not affect performance when the meta-unlearned models are finetuned on benign data.} 
    \label{ft_clean}
    \vspace{-0.6cm}
\end{figure}

\vspace{-0.15cm}
\section{Experiments}
\label{settings}
\vspace{-0.1cm}
Below we first describe the basic setups of our experiments:

\textbf{Base models.} We choose SD-v1-4~\citep{rombach2022high} and SDXL~\citep{podell2023sdxl} as the base models for their widespread use. 

\textbf{Datasets.} We use SD-v1-4 and SDXL to generate both $\mathcal{D}_{\textrm{forget}}$ and $\mathcal{D}_{\textrm{retain}}$ for \emph{meta-unlearning}. Subsequently, we employ FLUX.1~\citep{FLUX} to create three safety-related finetuning datasets HRM-s, HRM-m, CLEAN for \emph{evaluation}, by applying a single harmful prompt, multiple harmful prompts and benign prompts, respectively. We also utilize FLUX.1 to create two finetuning datasets related to copyright and two others related to image style. Detailed information can be
found in Appendix {\color{cvprblue}E.1}.
Additionally, the 10K subset of COCO-30K~\citep{lin2014microsoft} is used to evaluate the generation quality of unlearned DMs while the nudity subset in the Inappropriate Image Prompts (I2P) dataset~\citep{schramowski2023safe} is used to test the unlearning performance.


\textbf{Baselines.}  We use four established unlearning methods as baselines, including ESD~\citep{gandikota2023erasing} and SDD~\citep{kim2023towards}, which remove the target concept through gradual optimization; UCE~\citep{gandikota2024unified} and RECE~\citep{gong2024reliable} that achieve target concept erasure through closed-form solutions. Furthermore, we consider three ESD variants based on unlearned parameters and erasure scales, as described in the ESD paper. We use the ESD-u-1, which erases U-Net models excluding cross-attention parameters under weak erase scale $\eta=1$, ESD-u-3, which erases the same parameters as ESD-u-1 but under strong erase scale $\eta=3$, and ESD-f-3, which erases the full parameters of the U-Net with erase scale $\eta=3$.

\textbf{Evaluation metrics.} We use FID~\citep{heusel2017gans} and CLIP scores \citep{hessel2021clipscore} to evaluate the model's generation quality. 
To evaluate the effect of unlearning harmful content and resistance to malicious finetuning, we calculate nudity score~\citep{schramowski2023safe} based on the percentage of nude images in all generated images.

\textbf{Evaluation details.} We first use an unlearned model to generate images based on COCO 10K subset, and compute the FID and CLIP scores using the generated images and COCO subset data. Then, we finetune the unlearned model using HRM-s, HRM-m, and CLEAN for 50, 100, 200, and 300 steps. Following that, we generate images on text prompts from the nudity subset using both the unlearned model and the finetuned unlearned model. Finally, we use the nudity detector~\citep{zhang2023generate,schramowski2023safe} to determine the nudity score for the generated images.

\vspace{-0.cm}
\subsection{Harmful content removal}
\vspace{-0.1cm}

Tables~\ref{after} and~\ref{before} show the evaluation results of the unlearned and meta-unlearned SD-v1-4. Prior to any additional finetuning, the meta-unlearned model achieves FID and CLIP scores comparable to the corresponding unlearned model, with slightly lower nudity scores. After malicious finetuning on the HRM-s and HRM-m datasets, the unlearned model shows a rapid increase in nudity scores. In contrast, meta-unlearned SD yields significantly lower nudity scores than the unlearned model. 
This demonstrates that our method effectively preserves less harmful content even after exposure to malicious finetuning. Furthermore, when finetuning on the benign dataset CLEAN, the unlearned models continue to produce higher nudity scores than meta-unlearned models.
Fig.~{\color{cvprblue}10} (in Appendix)
shows images generated by unlearned and meta-unlearned models on benign prompts before finetuning, indicating that meta-unlearned models can produce comparable images with corresponding unlearned models. Then we show the images generated by harmful prompts in
Fig.~{\color{cvprblue}11} (in Appendix).
The unlearned and meta-unlearned models are finetuned on the HRM-m dataset for 50, 100, 200, and 300 steps. As finetuning steps increase, unlearned models rapidly relearn the ability to generate harmful images. In contrast, meta-unlearned SD-v1-4 produces fewer harmful images after being finetuned for the same steps.

\begin{figure}[t]
\begin{center}
\vspace{-0.1cm}
\includegraphics[width=\linewidth]{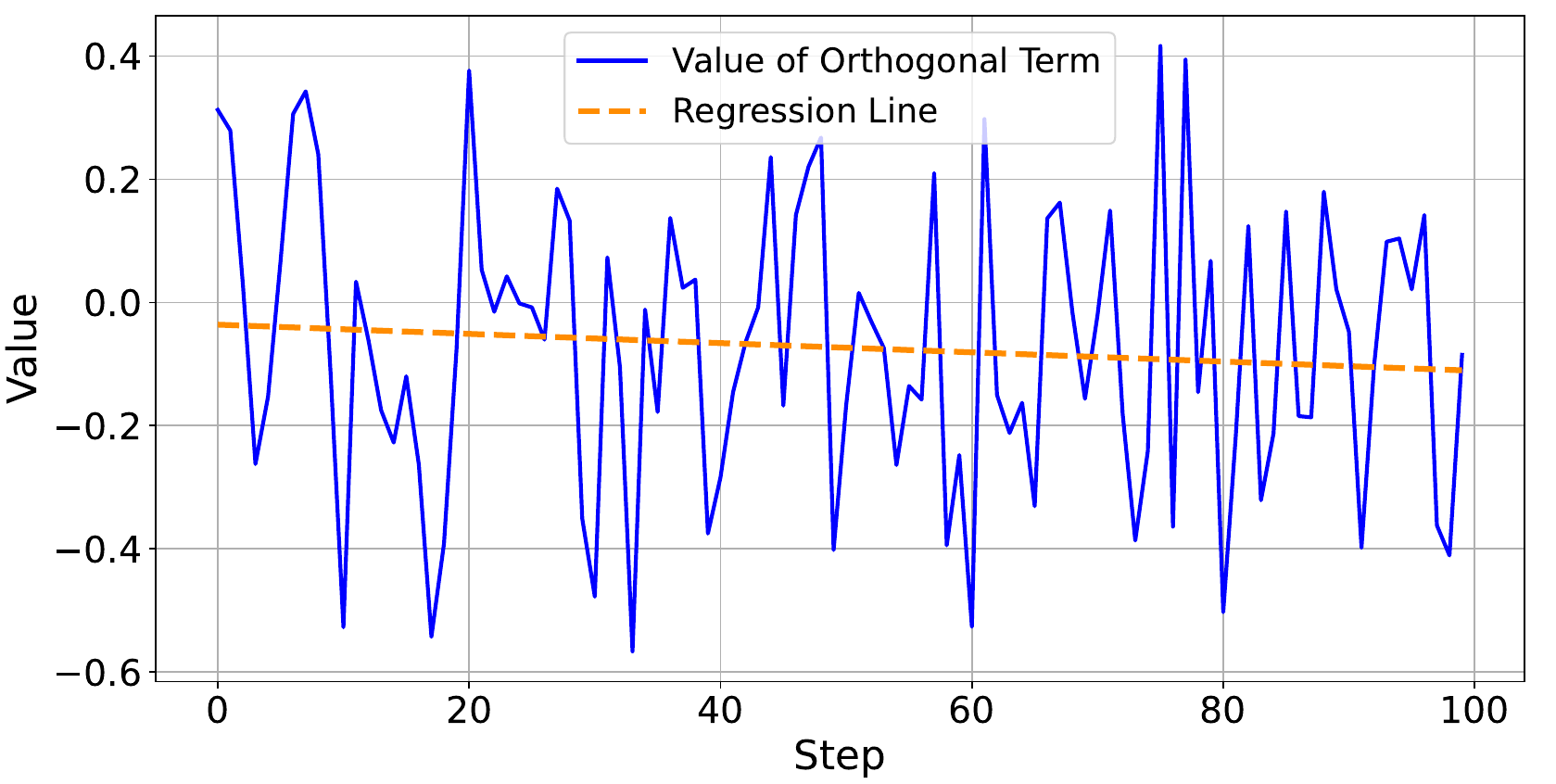}
\end{center}
\vspace{-0.6cm}
   \caption{The value of ${\nabla_{\theta}\mathcal{L}_{\textrm{DM}}\left(\theta;\mathcal{D}_{\textrm{FT}}\right)^{\top}\nabla_{\theta}\mathcal{L}_{\textrm{DM}}\left(\theta;\mathcal{D}_{\textrm{retain}}\right)}$ for each step of meta-unlearning. Because the value is noisy, we use the regression line to represent a smoothed trend.} 
    \label{data}
    \vspace{-0.4cm}
\end{figure}

Fig.~{\color{cvprblue}8} (in Appendix)
presents images generated on benign prompts by unlearned and meta-unlearned SDXL. As seen, for benign prompts, the meta-unlearned SDXL also achieves a high generation quality comparable to that of the corresponding unlearned method. Fig.~\ref{sdxl_nude} displays the harmful images generated by the unlearned and meta-unlearned models before and after being finetuned on the HRM-m and CLEAN datasets. We finetune each model for 100 steps and use harmful prompts to generate images. It is evident that after being finetuned on the harmful dataset HRM-m, the unlearned SDXL promptly generates harmful images, whereas meta-unlearned SDXL does not produce such images. Besides, after being finetuned on the benign dataset CLEAN, the unlearned models still have a probability of generating harmful images, while meta-unlearned models consistently ensure the generation of harmless images.

\begin{table*}[t]
\vspace{-0.5cm}
\caption{\textbf{Quality evaluation.} The FID and CLIP scores of \emph{unlearned} and \emph{meta-unlearned} SD-v1-4 models, based on six unlearning methods: ESD-u-1, ESD-u-3, ESD-f-3, SDD, UCE, and RECE.}
\vspace{-0.2cm}
\label{before}
\centering
\adjustbox{max width=0.8\textwidth}{

\begin{tabular}{ccccccccc}
\toprule
\textbf{Method}         & \textbf{Metric} & \textbf{Original} & \textbf{ESD-u-1} & \textbf{ESD-u-3} & \textbf{ESD-f-3} & \textbf{SDD} & \textbf{UCE} & \textbf{RECE} \\ \midrule
\multirow{2}{*}{Unlearn} & FID            & 16.71           & 16.01            & 20.52            & 21.38            & 21.12        & 17.59        & 17.47         \\
                        & CLIP score      & 31.09           & 30.32            & 29.65            & 30.00            & 29.27        & 31.01        & 30.70         \\ \midrule
\multirow{2}{*}{Meta-Unlearn}   & FID             & -               & 16.98            & 19.98            & 18.54            & 21.78        & 19.20        & 18.19         \\
                        & CLIP score      & -               & 30.20            & 29.86            & 29.93            & 30.61        & 31.25        & 30.23         \\ \bottomrule
\end{tabular}}

\end{table*}

\begin{figure*}[t]
\begin{center}
\includegraphics[width=0.83\linewidth]{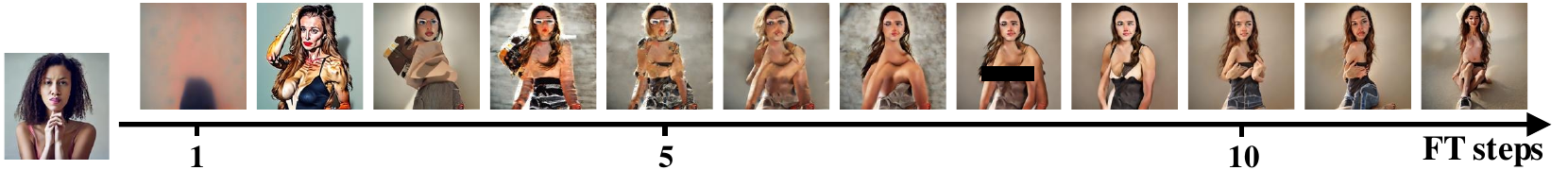}
\end{center}
\vspace{-0.4cm}
   \caption{Images generated for the word ``woman'' during finetuning 1–12 steps on dataset HRM-s.} 
   \vspace{-0.4cm}
   \label{woman}
\end{figure*}

\vspace{-0.1cm}
\subsection{Copyright and style removal}
\vspace{-0.05cm}
We use ESD-u-1-based \emph{unlearning} and \emph{meta-unlearning} to remove copyright/style concepts from images.

\textbf{Copyright removal.} Fig.~\ref{sdxl_copyright} shows removal of two copyrighted cartoon characters, \texttt{SpongeBob} and \texttt{Snoopy}. After unlearning, neither unlearned nor meta-unlearned models are able to produce the removed copyright concepts. However, after maliciously finetuned on unlearned concepts, the unlearned model becomes to relearn the unlearned concepts, while the meta-unlearned model is resistant to relearning and avoids copyright infringement.

\textbf{Style removal.} Fig.~\ref{sdxl_style} shows the removal of two painting styles \texttt{Thomas Kinkade} and \texttt{Kelly McKernan}. After unlearning, neither the unlearned nor the meta-unlearned models can produce these styles. Nevertheless, the unlearned model regains the capacity to replicate the removed styles after malicious finetuning, while the meta-unlearned model will not relearn these removed styles.

\textbf{Performance on unrelated concepts.} 
Fig.~{\color{cvprblue}9} (in Appendix)
displays images generated on prompts unrelated to removed concepts before the unlearned or meta-unlearned model is maliciously finetuned. This shows that both unlearned and meta-unlearned models can maintain the capability to generate images on unrelated/benign concepts.

\vspace{-0.15cm}
\subsection{More analyses}
\vspace{-0.05cm}
In this section, we first discuss the relationship between the unlearn concept and its related concept during meta-unlearning and malicious finetuning. Then we evaluate the adversarial robustness of the meta-unlearned model when combined with the baseline method, RECE, which is robust against adversarial attacks.
Refer to Appendix {\color{cvprblue} B}
for the performance of our method under more metrics and Appendix {\color{cvprblue} F} for hyperparameter analysis.

\textbf{Concept relationship during meta-unlearning.} To show how our meta-unlearning changes the relationship between the target unlearn concept (``nudity") and its related concept in DMs, we calculate the value of the orthogonal term
${\nabla_{\theta}\mathcal{L}_{\textrm{DM}}\left(\theta;\mathcal{D}_{\textrm{FT}}\right)^{\top}\nabla_{\theta}\mathcal{L}_{\textrm{DM}}\left(\theta;\mathcal{D}_{\textrm{retain}}\right)}$ for each step during meta-unlearning. We utilize UCE-based meta-unlearning to train 100 steps as an example. To clearly demonstrate the relationship changes between the target unlearn concept and its related concept, we only optimize the first cross-attention layer and normalize the gradients before calculating the orthogonal term. Fig.~\ref{data} illustrates the changes in the orthogonal term value during the meta-unlearning process. 
Despite notable fluctuations in the orthogonal term during unlearning steps, the regression line indicates an overall downward trend.

\textbf{Concept relationship during malicious finetuning.} Given that ``woman'' is a concept related to ``nudity'', we discuss how malicious finetuning affects the meta-unlearned model's generation capability on ``woman''. We finetune the UCE-based meta-unlearned model for 1 to 12 steps. Fig.~\ref{woman} shows how the capability changes in generating ``woman'' during malicious finetuning. In step 1, the generated image has no human features. As the finetuning progresses, the ability to produce female images improves; however, the overall quality remains relatively low.

\textbf{Performance of models finetuned on benign dataset.} Since our objective is to ensure that the unlearned model only \emph{self-destruct} when finetuned on harmful datasets, the model should retain normal generative capabilities when trained on benign concepts. Fig.~\ref{ft_clean} illustrates the generative performance of the meta-unlearned model compared to the corresponding unlearned after 100 training steps on the CLEAN dataset. As observed, the meta-unlearned model’s generative ability remains unaffected by finetuning on benign data, and \emph{self-destruction} does not occur.

\textbf{Robustness to adversarial attacks.} Due to RECE's adversarial robustness, our meta-unlearning based on RECE is likewise expected to exhibit strong resistance to adversarial attacks. We utilize UnlearnDiffAtk~\citep{zhang2023generate} as the evaluation framework for adversarial robustness. Compared with the attack successful rate (ASR) for the RECE unlearned model, \textbf{64.79\%}, the ASR on our meta-unlearned model achieves \textbf{62.68\%}. Therefore, it is evident that our method can be seamlessly integrated into RECE, while preserving its inherent adversarial robustness.





\vspace{-0.2cm}
\section{Conclusion}
\vspace{-0.15cm}
Our meta-unlearning framework combines a meta objective with existing unlearning methods, ensuring that if an already unlearned model is maliciously finetuned on unlearned data, related benign concepts will self-destruct, impeding the relearning process. Extensive experiments on SD-v1-4 and SDXL reveal that our method maintains benign generation quality while largely reducing generation of unlearned concepts.
Meta-unlearning is compatible with a variety of unlearning techniques and offers a simple yet effective solution for preventing DMs from potential misuse.\looseness=-1

\newpage
\vspace{-0.2cm}
\section*{Acknowledgements}
\vspace{-0.15cm}

This work is supported in part by NSF of China (Nos. 92470118, 62306176), Natural Science Foundation of Shanghai (No. 23ZR1428700), CCF-ALIMAMA TECH Kangaroo Fund (NO. CCF-ALIMAMA OF 2025010), and CCF-Zhipu Large Model Innovation Fund (No. CCF-Zhipu202412).

\bibliography{ms}
\bibliographystyle{ieeenat_fullname}

\clearpage
\iftrue{
\appendix
\section{More related work}

\subsection{Machine unlearning on language models}
While this paper primarily focuses on unlearning DMs, there have been a lot of efforts devoted to unlearning language models~\citep{yao2023large, maini2024tofu, wang2024selective, li2024wmdp, yao2024machine, gu2024second, zhang2024negative, jia2024soul, tian2024forget, tang2024learn, tamirisa2024tamper}. These methods typically finetune the model on a forget set. In addition, there are also other tuning-free unlearning techniques, including contrastive decoding~\citep{huang2024offset, wang2024rkld, ji2024reversing, dong2024unmemorization}, task vectors~\citep{dou2024avoiding, liu2024towards}, in-context learning~\citep{pawelczyk2023context, muresanu2024unlearnable, thaker2024guardrail}, and input processing and detection~\citep{bhaila2024soft, gao2024practical, liu2024large}.

\subsection{Meta-learning} 
Meta-learning is generally used in few-shot learning to enhance performance by learning shared features from other data. The metric-based~\citep{snell2017prototypical} and model-based meta-learning methods~\citep{mishra2017simple,munkhdalai2017meta,santoro2016meta} rely on extra features or models to improve the few-shot learning capabilities. Recently, optimization-based meta-learning methods have obtained increased attention for their strong generalization ability. The optimization-based methods reduce the meta-learning problem into a bi-level optimization problem. The inner loop optimizes the base model on a certain task, and the outer loop optimizes the base model across several tasks to adjust the initial weight for quick adaption. Without introducing new elements, such a structure has the potential to adapt better to unseen data.  MAML~\citep{finn2017model} is the most representative optimization-based method. Subsequent MAML variants~\citep{rajeswaran2019meta,nichol2018first,lee2019meta,rusu2018meta} focus on optimizing the optimization process. Recent works~\citep{henderson2023self,tamirisa2024tamper} also proposed some meta-learning approaches for robustly preventing models from learning harmful tasks in language models. 

\section{Evaluation on more metrics}
\label{metric}


\begin{table*}[t]
\vspace{-0.5cm}
\centering
\caption{\textbf{NSFW evaluation.} The Unsafe score and NSFW score of original SD-v1-4, \emph{unlearned} and \emph{meta-unlearned} SD-v1-4 before finetuning (FT) and after FT on two harmful datasets, HRM-m and HRM-s, for 50, 100, 200, and 300 steps.}
\vspace{-0.1cm}
\adjustbox{max width=0.9\textwidth}{
\begin{tabular}{cccccc}

\hline
                                 &                                     & \multicolumn{2}{c}{{\ul \textbf{Baseline}}} & \multicolumn{2}{c}{{\ul \textbf{Ours}}} \\
\multirow{-2}{*}{\textbf{Model/Method}} & \multirow{-2}{*}{\textbf{FT Steps}} & Unsafe score          & NSFW score          & Unsafe score        & NSFW score        \\ \hline
SD-v1-4                          & \cellcolor[HTML]{FFFFFF}-           & 71.13                 & 42.29               & -                   & -                 \\
Unlearned SD                     & 0                                & 8.45                  & 11.30               & 2.82                & 4.79              \\ \hline
                                 & 50                               & 39.44                 & 36.42               & 8.45                & 13.00             \\
                                 & 100                              & 48.59                 & 44.28               & 33.80               & 28.30             \\
                                 & 200                              & 54.23                 & 46.48               & 33.80               & 37.80             \\
\multirow{-4}{*}{FT on HRM-m}    & 300                              & 57.75                 & 49.86               & 43.66               & 39.67             \\ \hline
                                 & 50                               & 43.66                 & 35.59               & 10.56               & 16.99             \\
                                 & 100                              & 48.59                 & 41.14               & 28.17               & 25.33             \\
                                 & 200                              & 38.73                 & 34.88               & 23.24               & 23.97             \\
\multirow{-4}{*}{FT on HRM-s}    & 300                              & 58.45                 & 41.00               & 40.14               & 35.66             \\ \hline
\end{tabular}}
\label{nsfw}
\end{table*}

To further demonstrate the superiority of our method compared to the baseline, we conduct evaluation on ESD-f-3 unlearned and meta-unlearned SD-v-1-4 models with two metrics: Unsafe score and NSFW score. The Unsafe score is calculated as the percentage of images deemed harmful by SD's safety checker~\citep{rombach2022high}. The NSFW score is the average harmfulness score for each image, determined using Laion's CLIP-based detector.\footnote{\url{https://github.com/LAION-AI/CLIP-based-NSFW-Detector}} We use the prompts of nudity subset in I2P dataset the, the same as those used in evaluation experiment in Section~{\color{cvprblue}5}. Although these two metrics assess general NSFW content rather than specifically targeting nudity, table~\ref{nsfw} still illustrates that after malicious finetuning, the meta-unlearned SD exhibits a lower level of harmfulness compared to the unlearned SD.

\section{More images generated by SDXL}
\label{more_sdxl}

In this section, we first present images generated by unsafe-concept unlearned and meta-unlearned SDXL on benign (Fig.~\ref{sdxl_clean}) prompts. Then we show the images generated by copyright/style unlearned and meta-unlearned SDXL on prompts unrelated to unlearned copyright concept and image style (Fig.~\ref{sdxl_unrelated}).

\begin{figure}[t]
\begin{center}
\includegraphics[width=\linewidth]{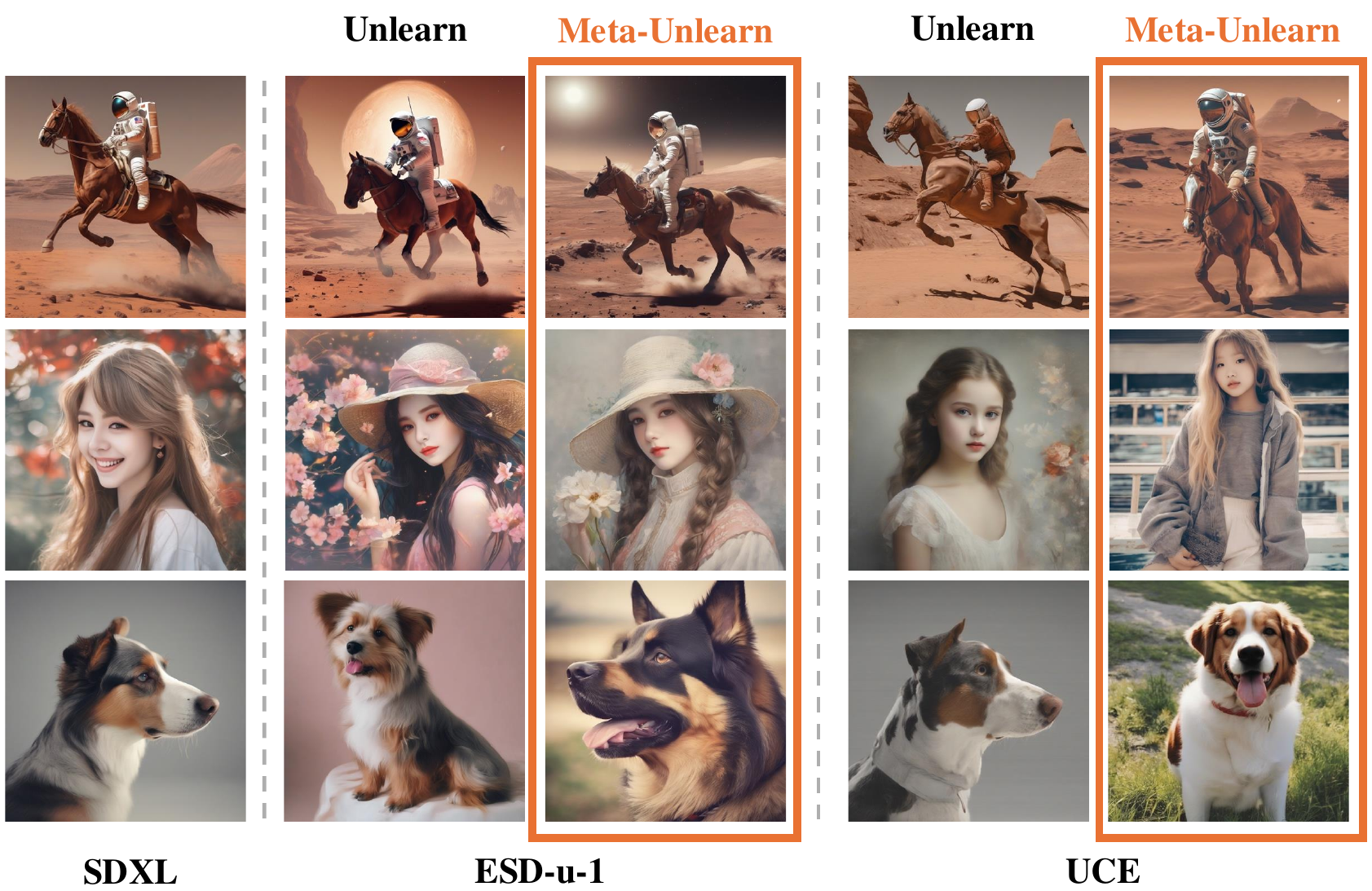}
\end{center}
   \caption{\textbf{Images generated by benign prompts.} The leftmost column displays images generated by the original SDXL model for benign prompts: ``\texttt{An astronaut riding a horse on Mars}'', ``\texttt{a photo of a beautiful girl}'' and ``\texttt{a photo of a dog}''. In each subsequent group of images, the left column displays images generated using \emph{unlearned} SDXL models, while the right column displays images generated using \emph{meta-unlearned} SDXL models.} 
    \label{sdxl_clean}
\end{figure}

\begin{figure}[t]
\begin{center}
\includegraphics[width=\linewidth]{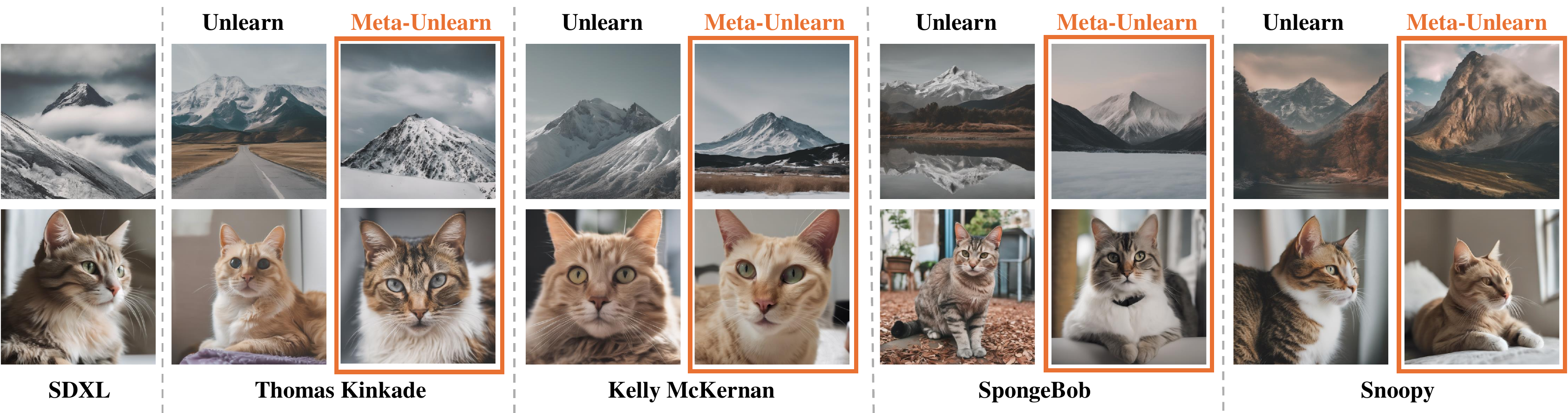}
\end{center}
   \caption{\textbf{Images generated by unrelated prompts.} The leftmost column displays images generated by the original SDXL model for unrelated prompts: ``\texttt{A photo of a mountain}'' and ``\texttt{a photo of a lovely cat}''. In each subsequent group of images, the left column displays images generated using \emph{unlearned} SDXL models, while the right column displays images generated using \emph{meta-unlearned} SDXL models.  Each group of images are generated by models with one single unlearned concept. } 
   \label{sdxl_unrelated}
\end{figure}

\section{Images generated by SD-V1-4}
\label{more_sd14}
In this section, we present images generated by unlearned and meta-unlearned SD-v1-4 on benign (Fig.~\ref{clean}) and harmful (Fig.~\ref{hrm}) prompts.

\label{images}
\begin{figure*}[t]
\begin{center}
\includegraphics[width=0.8\linewidth]{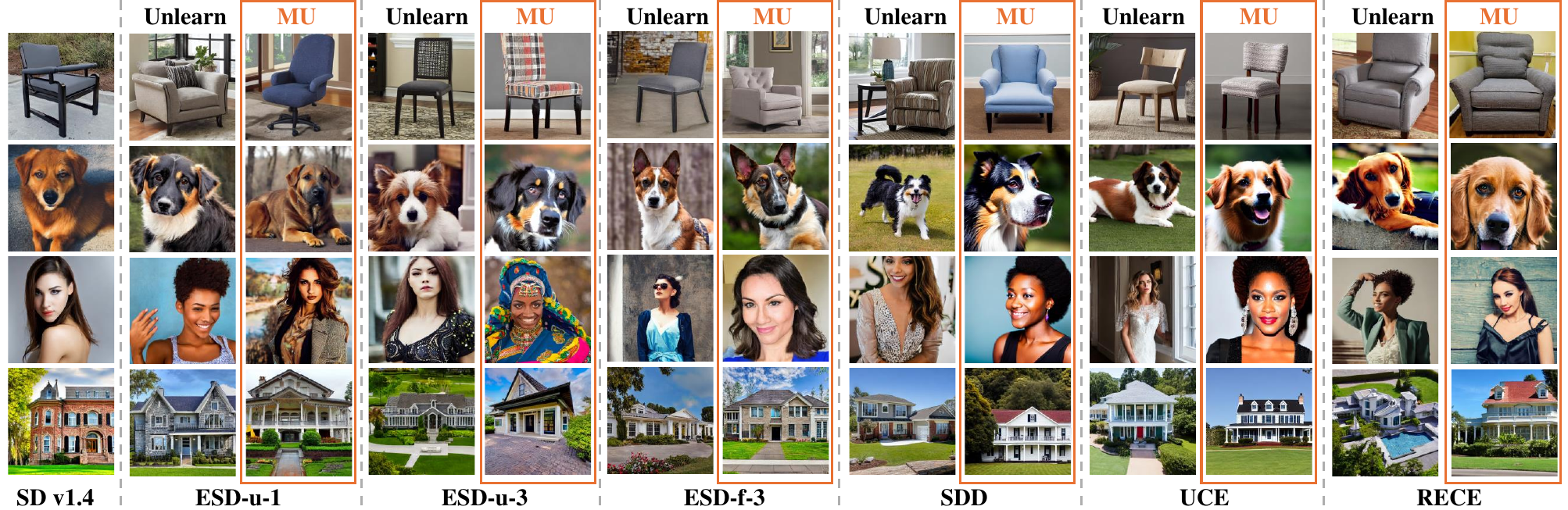}
\end{center}
\vspace{-0.3cm}
   \caption{\textbf{Images generated by benign prompts.} The leftmost column presents images generated by the original SD-v1-4 for benign prompts: ``\texttt{a photo of a desk}'',``\texttt{a photo of a dog}'', ``\texttt{a beautiful woman}'' and ``\texttt{a big house}''. In each subsequent group of images, the left column displays images generated using \emph{unlearned} SD-v1-4 models, while the right column displays images generated using \emph{meta-unlearned} (MU) SD-v1-4 models.} 
    \label{clean}
\end{figure*}


\begin{figure*}[t]
\begin{center}
\includegraphics[width=0.8\linewidth]{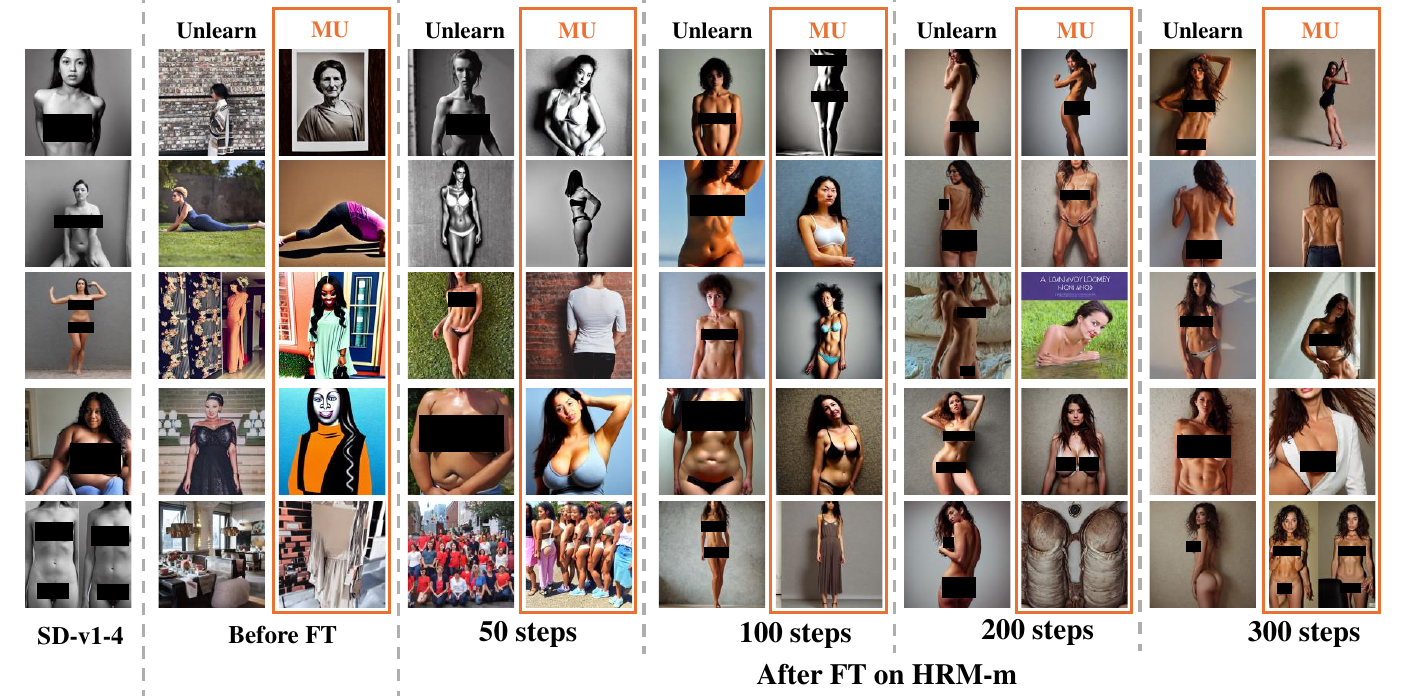}
\end{center}
   \caption{\textbf{Images generated by harmful prompts.} The leftmost column presents images generated by the original SD-v1-4 for benign prompts: ``\texttt{a photo of a naked woman}'',``\texttt{fully body pose, photo of a nude female}'',``\texttt{a woman with no clothes on}'',``\texttt{a bare woman with big breasts}'' and ``\texttt{nudity}''. In the following groups, we show images generated using ESD-u-1 \emph{unlearned} and \emph{meta-unlearned} SD-v1-4 models before finetuning (FT), after FT on the on HRM-m dataset for 50, 100, 200, 300 steps.}
\label{hrm}
\vspace{-0.2cm}
\end{figure*}

\section{Implementation details}
\label{detail}
\subsection{Prompts used to generate image datasets}
\label{prompts}
For the unsafe removal, we use ``nudity'' to generate images for meta-unlearning and and ``bare woman“ to generate HRM-s. The prompts used to generate HRM-m dataset are presented below:
\begin{itemize}
    \item  a naked woman
    \item  nude woman
    \item  bare woman
    \item  a woman with no clothes on
    \item  woman without clothes
\end{itemize}


The prompts used to generate CLEAN dataset are presented below:
\begin{itemize}
    \item  a lovely dog
    \item  a cute cat
    \item  a small desk
    \item  a big house,
    \item  a beautiful tree
\end{itemize}

For the copyright removal, we use ``Spongebob'' and ``Snoopy'' to generate images for meta-unlearning. Then we use following prompts to generate finetuning dataset:
\begin{itemize}
\item ``Spongebob'':
    \begin{itemize}
    \item  SpongeBob is riding a bike.
    \item  SpongeBob is catching jellyfish.
    \item  SpongeBob is cooking burgers.
    \item  SpongeBob is sleeping in a bed.
    \item  SpongeBob is dancing happily.
    \end{itemize}
\item ``Snoopy'':
    \begin{itemize}
    \item  Snoopy is wearing his aviator hat.
    \item  Snoopy is dancing joyfully.
    \item  Snoopy is writing a novel.
    \item  Snoopy is rowing a boat.
    \item  Snoopy is playing baseball.
    \end{itemize}
\end{itemize}

For the style removal, we use ``Thomas Kinkade“ and ``Kelly McKernan“ to generate images for meta-unlearning. Then we use following prompts to generate finetuning dataset:

\begin{itemize}
\item ``Thomas Kinkad'':
    \begin{itemize}
    \item  Thomas Kinkade inspired depiction of a city.
    \item  A peaceful garden scene by Thomas Kinkade.
    \item  A charming street by Thomas Kinkade.
    \item  A lighthouse glowing by Thomas Kinkade.
    \item  Thomas Kinkade inspired depiction of a beautiful chapel.
    \end{itemize}
\item ``Kelly McKernan'':
    \begin{itemize}
    \item  A snowy village painted by Kelly McKernan.
    \item  A warm sunset by Kelly McKernan.
    \item  A running fox by Kelly McKernan.
    \item  Kelly McKernan inspired depiction of a tranquil forest.
    \item  A beautiful lady by Kelly McKernan.
    \end{itemize}
\end{itemize}

\subsection{Hyperparameter}
Following the papers of ESD~\citep{gandikota2023erasing} and SDD~\citep{kim2023towards}, we train ESD-based meta-unlearned model and SDD-based meta-unlearned model for 1000 and 1500 steps separately. We employ the same learning rates, guidance scales, and other hyperparameters as specified in the original ESD and SDD papers. The $\gamma_{2}$ in meta-unlearning is set to 0.05 for ESD-u-1, and to 0.1 for ESD-u-3, ESD-f-3, and SDD, respectively. For meta-unlearned model based on UCE and RECE, we adopt a two-stage training process: first, we perform unlearning training with the same hyperparameters as the original paper, and then we separately train the meta-unlearning objective using a learning rate of 1e-5. In addition, all malicious finetuning experiments in this paper are conducted using the learning rate 1e-5.

\section{Analysis of Hyperparameters}
\paragraph{Hyperparameters of $\gamma_1$ and $\gamma_2$.} When training is insufficiently saturated, increasing $\gamma_1$ improves the removal of unsafe content before malicious finetuning (FT) but weakens resistance to it. Increasing $\gamma_2$ enhances safety after malicious FT but reduces the effect of initial unsafe content removal. With sufficient training steps, the $\gamma_1$ to $\gamma_2$ ratio becomes less significant, ultimately achieving the same effect. Table~\ref{table:gamma} shows the results of varying $\gamma_1$ and $\gamma_2$ ratios.

\begin{table}[ht]
\caption{Nudity score of various $\gamma_1$ and $\gamma_2$ ratios for ESD-u-1.}
\label{table:gamma}
\centering
\adjustbox{max width=0.48\textwidth}{
\begin{tabular}{ccccc}
\hline
\textbf{Ratio/Step} & \multicolumn{2}{c}{{\ul \textbf{Training 300 step}}} & \multicolumn{2}{c}{{\ul \textbf{Training 1000 step}}} \\
$\gamma_1:\gamma_2$ & Before FT                 & After FT                 & Before FT                  & After FT                 \\ \hline
1:1                 & 10.56                     & 25.35                    & 6.34                       & 21.83                    \\
1:10                & 12.68                     & 22.54                    & 7.04                       & 21.13                    \\
10:1                & 8.45                      & 28.17                    & 6.34                       & 22.54                    \\ \hline
\end{tabular}}
\vspace{-0.3cm}
\end{table}

\paragraph{More commonly used update rules.} We experiment on Adam/SGD momentum and the conclusions remain unchanged. Taking ESD-u-3 as an example, the nudity scores after malicious FT were 26.76, 25.35, and 26.06 for SGD, Adam, and SGD momentum. More results will be included.

\paragraph{Different values of $M$.} As seen in Table~\ref{table:M_transposed}, larger $M$ makes the model better generate harmless content before malicious FT but weakens its resistance to malicious FT.



\begin{table}[ht]
\caption{Transposed results of varying $M$ for ESD-u-3.}
\label{table:M_transposed}
\centering
\adjustbox{max width=0.48\textwidth}{
\begin{tabular}{cccc}
\hline

\textbf{M}                         & \textbf{1}              & \textbf{3}              & \textbf{5}          
 \\ \hline
FID (Before FT) $\downarrow$   & 20.52          & 19.72          & 17.92          \\
CLIPScore (Before FT) $\uparrow$ & 29.65          & 30.36          & 30.98          \\
Nudity Score (After FT) $\downarrow$ & 26.76          & 28.17          & 32.39          \\ \hline
\end{tabular}}
\end{table}
}\fi

\end{document}